\documentclass[letterpaper, 10 pt, conference]{ieeeconf}  

\IEEEoverridecommandlockouts                              

\usepackage{multirow}
\usepackage{amssymb}
\usepackage{xcolor}
\usepackage[utf8]{inputenc}
\usepackage{color}
\usepackage{graphicx}
\usepackage{float}
\usepackage{comment}
\usepackage{placeins}

\usepackage{fancyhdr}

\date{June 2019}

\begin{document}

\title{\LARGE \bf Unsupervised Pre-trained Models from Healthy ADLs Improve Parkinson's Disease Classification of Gait Patterns}

\author{Anirudh Som\textsuperscript{1,2}, Narayanan Krishnamurthi\textsuperscript{3}, Matthew Buman\textsuperscript{4}, Pavan Turaga\textsuperscript{1,2}}







\maketitle

\begin{abstract}
     
     \textit {Application and use of deep learning algorithms for different healthcare applications is gaining interest at a steady pace. However, use of such algorithms can prove to be challenging as they require large amounts of training data that capture different possible variations. This makes it difficult to use them in a clinical setting since in most health applications researchers often have to work with limited data. Less data can cause the deep learning model to over-fit. In this paper, we ask how can we use data from a different environment, different use-case, with widely differing data distributions. We exemplify this use case by using single-sensor accelerometer data from healthy subjects performing activities of daily living - ADLs (source dataset), to extract features relevant to multi-sensor accelerometer gait data (target dataset) for Parkinson's disease classification. We train the pre-trained model using the source dataset and use it as a feature extractor. We show that the features extracted for the target dataset can be used to train an effective classification model. Our pre-trained source model consists of a convolutional autoencoder, and the target classification model is a simple multi-layer perceptron model. We explore two different pre-trained source models, trained using different activity groups, and analyze the influence the choice of pre-trained model has over the task of Parkinson's disease classification.}
     
\end{abstract}

\let\thefootnote\relax\footnotetext{\vspace{-0.2in}\\
\textsuperscript{1}School of Arts, Media and Engineering, Arizona State University (ASU)\\
\textsuperscript{2}School of Electrical, Computer and Energy Engineering, ASU\\
\textsuperscript{3}Edson College of Nursing and Health Innovation, ASU\\
\textsuperscript{4}College of Health Solutions, ASU\\
\indent This work was supported in part by NIH R01GM135927, NSF CAREER grant number 1452163 and by the National Institute of Child Health and Human Development, NIH 1R21HD060315. ASU’s institutional review board approved all study materials and procedures (protocol number 1304009121 and protocol number 0808003170).}

\vspace{-0.015in}
\section{Introduction and Related Work}\label{section_intro}
\vspace{-0.025in}
Recent advances in wearable technologies like smart-watches and fitness trackers has proven to be an accessible and low-cost approach for a variety of activity-based health interventions. These devices contain inertial measurement unit (IMU) sensors like accelerometers, gyroscopes that help monitor movements continuously for extended periods during daily activities. Data from these devices together with sophisticated machine learning algorithms like deep learning can help characterize human movement and develop automated systems for many applications in movement disorders such as Parkinson's \cite{hammerla2015pd,alsheikh2016deep,cheng2017human,mohammadian2018novelty,san2019increasing} and human activity recognition for health and well-being interventions \cite{wang2019deep,west2015strava,sazonov2015posture,lu2017towards}. With the rise in deep learning algorithms, hand-engineered features have been replaced by features learnt by data-driven methods. However, for robust performance, they require a substantial amount of data to do well at inference. Gaining access to large amount of clinically relevant movement data, can be difficult, expensive and can lead to privacy-related issues. One can address this issue by using pre-trained models trained using a larger source dataset. Part of the pre-trained model can be used as a feature extraction tool for the target dataset of interest. The features extracted are later used to train a smaller, simpler classification model. However, this technique assumes that the source and target datasets have similar data distributions and data collection environments, \emph{i.e.}, same sensor-device, data collection protocol, sensor placement on the body, \emph{etc.} This assumption is rarely applicable to real-world applications. An unsupervised pre-trained model can help address this issue to a certain extent as it learns to characterize data without taking the associated class labels into account. 

In this paper, we ask whether movement data acquired from wearable devices for one specific intervention can be used to learn deep-learning models, but applied to an entirely different end-use robustly. To address this question, we use two specific situations. For the source domain, we assume access to accelerometry data from general health and well-being interventions, including tracking of activities of daily living. We seek to apply features learnt from the source to the target domain of accelerometer-based Parkinson's disease gait-based assessment. The motivation for this is that while general purpose activities of daily living can be obtained relatively easily, including from public databases like USC-HAD \cite{zhang2012usc}, it is much harder to obtain large-scale gait-data from special populations like Parkinson's disease.

Parkinson's disease is the second most common neurodegenerative disease in the world \cite{reeve2014ageing}. Symptoms include postural instability, gait dysfunction, speech degradation, motor function impairment, erratic behavior and thought process. It is estimated that about one million people are afflicted by the disease in the United States alone and live with no cure \cite{nih2015}. The most common approach to detect presence of Parkinson's disease consists of questionnaires and visual evaluation of disease-specific impairments by a clinical expert. However, these evaluations can be prone to subjective bias. An ideal scenario would involve a consensus evaluation by multiple clinicians, but this would result in being an expensive and time-consuming process for the patient. 

In this paper we use an unsupervised pre-trained model (trained using a source dataset containing single-sensor ADLs data from healthy subjects) as a feature extractor for the target dataset of interest. Here, the target dataset consists of multi-sensor gait data. We use the extracted features for the task of binary classification of gait patterns into healthy or Parkinson's disease subjects. Note, the source and target datasets share no similarity. We also explore the influence the distribution of the different classes in the source dataset has on the final binary classification task. Variants of the proposed approach have been successfully applied to other clinical and non-clinical applications \cite{cabrera2017automatic,malhotra2017timenet,freitag2017audeep,lyu2018improving,gupta2018using,narejo2016eeg}. The rest of the paper is organized as follows -- Section \ref{section_learning_strategies} provides background information on supervised and unsupervised learning. Section \ref{section_network_architecture} goes over details of the autoencoder and classification model used. Section \ref{section_dataset} gives a detailed description of the source and target datasets. In Section \ref{section_experiments} we discuss the binary classification experiment results. Section \ref{section_conclusion} concludes the paper. 


\vspace{-0.015in}
\section{Background}\label{section_learning_strategies}
\vspace{-0.025in}

\textbf{Supervised learning} is concerned with learning complex mappings from $X$ to $Y$ when many pairs of $(x,y)$ are given as training data. Here, $x\in X$ and $y\in Y$ are the input and output variables respectively. In a classification setting $Y$ corresponds to a fixed set of labels. On the other hand, \textbf{unsupervised learning} algorithms assume not having access to the label information of the data samples, thereby allowing us to learn the underlying patterns and characteristics of the data without making any assumptions of the associated class labels. An \textbf{autoencoder} is a popular unsupervised learning algorithm. It focuses on learning mappings from $X$ to $X$, \emph{i.e.}, the output of the model is set equal to the model's input. In other words, it tries to behave like an identity function. An autoencoder consists of two parts: (1) Encoder, (2) Decoder. At the time of training, the encoder learns to map the input data to a latent space representation, while the decoder learns to map the projected latent representation to the output of the model. At inference time, if $x$ is passed as input then $\hat x$ is obtained as output, with $\hat x$ being very similar to $x$. The mean-squared-error loss function is used to update the model's weights. Using a \textbf{pre-trained autoencoder} -- trained using a source dataset ($\mathcal{D}_s$), we can compute latent representations of the target training dataset ($\mathcal{D}_t$). The projected latent representations of $\mathcal{D}_t$ can be used to train a new classifier for the target classification task ($\mathcal{T}_t$).

Training deep learning models in a supervised fashion is suitable only when there is a large amount of training data that captures different variations. Often these models are trained using clean, uniformly distributed source datasets that are collected in well-defined controlled environments. They assume that the target data of interest is also collected in a similar environment and adheres to the same distribution as the source dataset. However, this is never the case and collecting vast quantities of data in a healthcare setting can prove to be a challenge. Also, training a deep learning model with limited amount of data can cause the model to over-fit. Data augmentation and domain adaptation techniques have been employed to handle these issues but mainly for visual classification tasks \cite{wang2018deep,shorten2019survey}. It would be difficult to apply these techniques in a healthcare setting, especially for time-series data, where the data environment continuously changes. Due to this reason we explore using pre-trained unsupervised autoencoder models for feature extraction. The autoencoder is trained using a larger $\mathcal{D}_s$ -- comprised of different activities performed by healthy subjects. It is later used to extract latent feature representations for a smaller $\mathcal{D}_t$ -- consisting of gait data from healthy and Parkinson's disease subjects. 

\vspace{-0.015in}
\section{Network Architecture}\label{section_network_architecture}
\vspace{-0.025in}

Here we go over the network architecture and hyper-parameter settings for the source autoencoder model and the target classification model. 

\vspace{-0.025in}
\subsection{Autoencoder Model}
We use a temporal \emph{DenseNet} architecture \cite{huang2017densely} to build the autoencoder model, with the \emph{DenseNet} model being a variant of Convolutional Neural Networks (CNNs). There have been previous works that explored the use of Recurrent Neural Networks (RNNs) instead for Parkinson's disease modeling \cite{che2017rnn}. However, a recent study suggests that temporal CNNs have a longer memory retention capacity and outperform RNNs on a diverse range of tasks and datasets \cite{bai2018empirical}. For this reason we use the \emph{DenseNet} architecture for building the autoencoder model. We set the number of dense blocks in the encoder and the decoder to 2. The following hyper-parameter settings were used: number of layers per dense block = 4, bottleneck size = 4, initial number of convolution filters = 32, initial convolution filter width = 7, initial pool width = 3, number of convolution filters = 16, convolution filter width = 3, transition pool size = 2, stride = 1, theta = 0.5, dropout rate = 0.2. We set stride = 1 as this helps keep the temporal dimension of the input signal unchanged throughout the autoencoder. The autoencoder model was used only to train on the source dataset in our experiments. The total number of trainable parameters is 264265.

\vspace{-0.025in}
\subsection{Multi Layer Perceptron (MLP) Model}
The MLP model was used as the target classification model. It contains 4 dense layers with ReLU activation and having 64, 128, 128, 64 units respectively. To avoid over-fitting, each dense layer is L2 regularized and followed by a dropout layer with a dropout rate of 0.2. The output layer is another dense layer with Softmax activation and with number of units equal to the number of classes. The total number of trainable parameters is a little over 35000, which is still a lot less compared to the pre-trained autoencoder model.

\vspace{-0.015in}
\section{Dataset}\label{section_dataset}
\vspace{-0.025in}


\vspace{-0.025in}
\subsection{Source Dataset}
\begin{figure}[t!]
	\centering
	\vspace{0.075in}
	\includegraphics[width=0.875\linewidth]{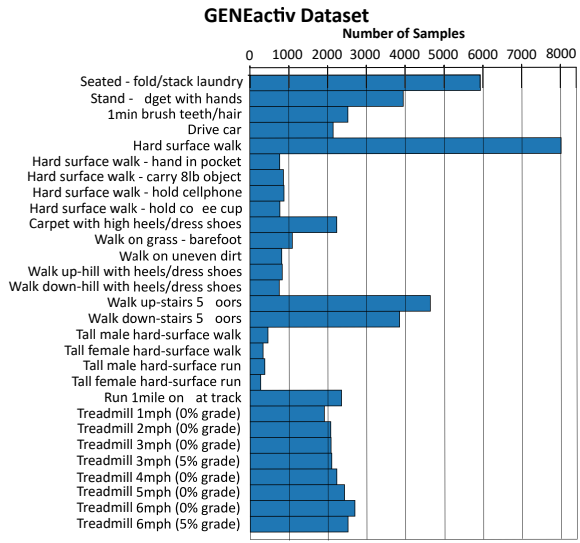}
	\vspace{-0.1in}
	\caption{Distribution of activity classes in the source dataset, collected using the \emph{GENEactiv} sensor \cite{wang2016statistical}.} \label{geneactiv_distribution}
	\vspace{-0.25in}
\end{figure}

The source dataset consists of 29 different activity classes from 152 healthy subjects. It was collected using the \emph{GENEactiv} sensor, a single wrist worn accelerometer sensor at a sampling rate of 100Hz. Figure \ref{geneactiv_distribution} shows the distribution of the different activity classes. Detailed description of the subject characteristics and data collection protocol can be found here \cite{wang2016statistical}. In this dataset, we considered two different subsets with eight activities each, to serve as the source dataset in our experiments. This was done to check if the type of activities present in the source dataset influenced the target binary classification task in any way. 

\noindent\textbf{Subset-1:} Contains treadmill activities (\emph{i.e., primarily walking}) performed at different speeds and inclincations -- Treadmill 1mph (0\% grade), Treadmill 2mph (0\% grade), Treadmill 3mph (0\% grade), Treadmill 3mph (5\% grade), Treadmill 4mph (0\% grade), Treadmill 5mph (0\% grade), Treadmill 6mph (0\% grade), Treadmill 6mph (5\% grade).

\noindent\textbf{Subset-2:} Contains four non-ambulatory and four treadmill activities -- Seated-folding/stacking laundry, Standing/fidgeting with hands, 1min brush teeth/1min brush hair, Driving car, Treadmill 1mph (0\% grade), Treadmill 3mph (0\% grade), Treadmill 5mph (0\% grade), Treadmill 6mph (5\% grade).

\vspace{-0.025in}
\subsection{Target Dataset}

\noindent\textbf{Subject Characteristics and Selection Criteria:}  The target gait dataset consists of 16 healthy and 18 Parkinson's disease (PD) subjects. Age of healthy subjects ranged from 52 - 75. PD subjects were selected if they satisfied the following conditions: PD diagnoses is in accordance with the UK Brain Bank criteria; are aged between 30 - 80; have a Hoehn-Yahr score between 2 and 3.5 (on a scale of 0 to 5) during medication-off/Deep Brain Stimulation-on condition; are able to participate in walking and standing trials without assistance; are at least three months post-implantation of Deep Brain Stimulation (DBS) device(s) (unilateral or bilateral); have stable stimulator settings and an antiparkinsonian medication regime (as judged by the screening clinician) for at least two weeks before their experimental evaluation visit. Individuals with PD exhibiting any of the following conditions were excluded from the study: have a recent history of unstable heart or lung disease; have evidence of pregnancy; have a history of non-compliance with medical or research procedures; have untreated chemical addiction or abuse; have an uncontrolled psychiatric illness; have major neurological (\emph{e.g.}, stroke), musculoskeletal (\emph{e.g.}, rheumatoid arthritis), or metabolic (\emph{e.g.}, diabetes) problems; have cognitive impairment (score of less than 25 in the mini-mental state examination); are unable to walk or stand without any walking aid (\emph{e.g.}, using a cane) for any reason; and presence of significant dyskinesia. 

\noindent\textbf{Subject Evaluation and Gait Data Collection:} Gait and severity of PD symptoms were evaluated in the medication-off condition at three different DBS frequency settings: (1) clinically determined setting (CDS); (2) intermediate frequency (INT) setting, where the frequency was reduced to about 80Hz; and (3) low frequency (LOW) setting, with the frequency further reduced to about 30Hz. During INT and LOW conditions only the frequency of the stimulation was altered from that of the CDS condition, with all other parameters such as stimulation amplitude, pulse width, \emph{etc.} being unchanged. Note, PD subjects had to discontinue antiparkinsonian medications at least 12 hours before participating in the clinical evaluations.

To assess gait, both PD and healthy subjects were asked to wear six small, light-weight sensors in the following regions: \emph{Sternum, Lumbar, Left-Ankle, Right-Ankle, Left-Wrist, Right-Wrist}. Acccelerometer data was collected at a frequency of 128Hz. These sensors were connected to a data logger that the subjects wore. The setup did not affect a subject's walking patterns. The gait protocol consisted of walking along a 30 meter straight path, turning around, and continuing to walk along the same path. Each subject carried out this protocol 1-2 times in each trial. Note, for all subjects the gait trials were collected in all three frequency settings with the first setting always being CDS. However, the order of data collection in INT and LOW settings was randomized.

\vspace{-0.015in}
\section{Experiments}\label{section_experiments}
\vspace{-0.025in}

\begin{table*}[h!]
	\centering
	\vspace{0.075in}
	\scalebox{0.735}{
	\begin{tabular}{ |c|c|c|c|c|c|c|c|c|c|c|c|c|c| } 
			\hline
			\textbf{Joint} &
			\textbf{Subject}  & \textbf{RMS$\mathbf{(X)}$} & \textbf{RMS$\mathbf{(Y)}$} & \textbf{RMS$\mathbf{(Z)}$} & $\mathbf{\rho(X,Y)}$ & $\mathbf{\rho(Y,Z)}$ & $\mathbf{\rho(X,Z)}$ & $\mathbf{dx}$ & $\mathbf{dy}$ & $\mathbf{dz}$ \\
			\hline
			\multirow{2}{*}{Sternum} & Healthy  & 0.9959$\pm$0.0029 & 0.9734$\pm$0.0121  & 0.9580$\pm$0.0171  & -0.0119$\pm$0.1127  & -0.2445$\pm$0.2864  & 0.0533$\pm$0.1318  & 4.5800$\pm$0.4359  & 5.2700$\pm$0.5642  &  5.2720$\pm$0.4327   \\
			\cline{2-11}  & PD  & 0.9811$\pm$0.0249  & 0.9282$\pm$0.0421  & 0.9066$\pm$0.0708  & 0.0222$\pm$0.1586  & -0.1043$\pm$0.4224  & -0.0180$\pm$0.1408  & 5.0328$\pm$0.7466  & 5.3082$\pm$0.7351  & 5.2587$\pm$0.6879    \\
			
			\hline
			\multirow{2}{*}{Lumbar} & Healthy  & 0.9309$\pm$0.2424  & 0.8998$\pm$0.2347  & 0.9174$\pm$0.2389  & -0.0624$\pm$0.0919  & -0.2760$\pm$0.2084  & 0.0035$\pm$0.1271  & 4.4058$\pm$1.2355  & 6.4199$\pm$2.0201  &  5.1923$\pm$1.6112   \\
			\cline{2-11}  & PD  & 0.9889$\pm$0.0125  & 0.9421$\pm$0.0228  & 0.9538$\pm$0.0231  & -0.0193$\pm$0.1315  & -0.4891$\pm$0.2733  & 0.0659$\pm$0.1884  & 5.2806$\pm$0.9654  & 6.4865$\pm$1.3118  &  5.3376$\pm$0.7821   \\
			
			\hline
			\multirow{2}{*}{Left-Ankle} & Healthy  & 0.9888$\pm$0.0057  & 0.9523$\pm$0.0314  & 0.9845$\pm$0.0081  & 0.0887$\pm$0.1452  & 0.1609$\pm$0.1150  & 0.3475$\pm$0.4597  & 5.0516$\pm$0.2947  & 7.9220$\pm$1.9291  &  6.4418$\pm$0.8733   \\
			\cline{2-11}  & PD  & 0.9794$\pm$0.0292  & 0.9409$\pm$0.0506  & 0.9783$\pm$0.0203  & -0.0313$\pm$0.2474  & 0.2257$\pm$0.1196  & -0.0006$\pm$0.5899  & 5.2021$\pm$0.6200  & 7.8417$\pm$1.5908  & 7.0807$\pm$1.3558    \\
			
			\hline
			\multirow{2}{*}{Right-Ankle} & Healthy  & 0.9890$\pm$0.0052 & 0.9650$\pm$0.0210  & 0.9851$\pm$0.0080  & 0.1678$\pm$0.1544  & 0.1976$\pm$0.1189  & 0.6554$\pm$0.2881  & 5.0975$\pm$0.3621  & 6.7725$\pm$0.8191  & 6.6457$\pm$0.7158   \\
			\cline{2-11}  & PD  & 0.9813$\pm$0.0250  & 0.9390$\pm$0.0513  & 0.9801$\pm$0.0182  & 0.2076$\pm$0.1967  & 0.2438$\pm$0.1627  & 0.4763$\pm$0.4301  & 5.1973$\pm$0.5391  & 7.4826$\pm$1.3098  & 7.3324$\pm$1.3710     \\
			
			\hline
			\multirow{2}{*}{Left-Wrist} & Healthy  & 0.9547$\pm$0.0265  & 0.8664$\pm$0.0716  & 0.8633$\pm$0.0534  & 0.1702$\pm$0.1727  & -0.2037$\pm$0.2186  & -0.2039$\pm$0.2848  & 5.9410$\pm$1.7871  & 7.5391$\pm$2.6950  & 6.2954$\pm$2.1241    \\
			\cline{2-11}  & PD  & 0.9257$\pm$0.0706  & 0.7908$\pm$0.1457  & 0.8199$\pm$0.1138  & 0.2946$\pm$0.2319  & -0.0955$\pm$0.2312  & -0.2093$\pm$0.3066  & 5.0623$\pm$0.8364  & 5.4300$\pm$1.3751  & 4.7666$\pm$1.2171    \\
			
			\hline
			\multirow{2}{*}{Right-Wrist} & Healthy  & 0.9476$\pm$0.0280  & 0.8757$\pm$0.0743  & 0.8669$\pm$0.0573  & -0.0531$\pm$0.2164  & -0.1664$\pm$0.1836  & 0.3840$\pm$0.2518  & 5.6958$\pm$1.3116  & 6.9585$\pm$2.3366  & 5.7788$\pm$1.4242   \\
			\cline{2-11}  & PD  & 0.9158$\pm$0.0747  & 0.7982$\pm$0.1366  & 0.8079$\pm$0.1147  & -0.1424$\pm$0.2289  & -0.0318$\pm$0.1931  & 0.4056$\pm$0.2823  & 4.9068$\pm$1.0396  & 4.9164$\pm$1.0992  & 4.4386$\pm$1.1866   \\

			\hline
	\end{tabular}}
	\caption{Mean $\pm$ Std of different statistical summaries for each joint, computed across different trials performed by Healthy and Parkinson's disease subjects.}\label{joint_statistics_info}
\end{table*}

\begin{table*}[h!]
	\centering
	\vspace{-0.025in}
	\scalebox{0.925}{
	\begin{tabular}{ |c|c|c|c|c|c| } 
			\hline
			\textbf{Classifier} &
			\textbf{Feature Representation} & \textbf{Accuracy} & \textbf{Precision} & \textbf{Recall} & \textbf{F1-Score} \\
			\hline
			\multirow{6}{*}{SVM} &
			Hand-engineered Features \cite{wang2016statistical} & 56.73$\pm$5.50 & 58.34$\pm$7.35 & 56.73$\pm$5.50 & 56.65$\pm$6.08 \\
			\cline{2-6}  &
			Time-series & 50.17$\pm$1.42 & 51.44$\pm$2.62 & 50.17$\pm$1.42 & 50.51$\pm$1.68 \\
			\cline{2-6} &
			\textbf{Pretrained Autoencoder - Unconstrained Latent Representations (Subset-1)} & \textbf{67.72$\pm$1.46} & \textbf{72.25$\pm$2.79} & \textbf{67.72$\pm$1.46} & \textbf{67.67$\pm$1.69} \\
			\cline{2-6}
			  &
			\textbf{Pretrained Autoencoder - Unconstrained Latent Representations (Subset-2)} & \textbf{66.15$\pm$5.72} & \textbf{69.54$\pm$6.00} & \textbf{66.15$\pm$5.72} & \textbf{66.19$\pm$5.97} \\
			\cline{2-6} &
			\textbf{Pretrained Autoencoder - Constrained Latent Representations (Subset-1)} & \textbf{68.92$\pm$4.42} & \textbf{72.72$\pm$5.63} & \textbf{68.92$\pm$4.42} & \textbf{68.75$\pm$4.14} \\
			\cline{2-6}
			  &
			\textbf{Pretrained Autoencoder - Constrained Latent Representations (Subset-2)} & \textbf{68.20$\pm$8.33} & \textbf{72.67$\pm$8.48} & \textbf{68.20$\pm$8.33} & \textbf{68.16$\pm$8.24} \\
			\hline
			
			\cline{1-6}
			\multirow{6}{*}{MLP} &
			Hand-engineered Features  \cite{wang2016statistical} & 61.60$\pm$1.81 & 63.62$\pm$2.32 & 61.60$\pm$1.81 & 61.30$\pm$2.26 \\
			\cline{2-6}  &
			Time-series & 60.57$\pm$5.96 & 61.22$\pm$5.74 & 60.57$\pm$5.95 & 60.00$\pm$5.23 \\
			\cline{2-6} &
			\textbf{Pretrained Autoencoder - Unconstrained Latent Representations (Subset-1)} & \textbf{69.13$\pm$4.94} & \textbf{70.46$\pm$4.66} & \textbf{69.13$\pm$4.94} & \textbf{68.83$\pm$4.44} \\
			\cline{2-6}
			  &
			\textbf{Pretrained Autoencoder - Unconstrained Latent Representations (Subset-2)} & \textbf{68.64$\pm$1.28} & \textbf{69.27$\pm$2.64} & \textbf{68.64$\pm$1.28} & \textbf{67.65$\pm$1.53} \\
			\cline{2-6} &
			\textbf{Pretrained Autoencoder - Constrained Latent Representations (Subset-1)} & \textbf{73.81$\pm$5.88} & \textbf{76.53$\pm$5.78} & \textbf{73.81$\pm$5.88} & \textbf{73.89$\pm$5.69} \\
			\cline{2-6}
			  &
			\textbf{Pretrained Autoencoder - Constrained Latent Representations (Subset-2)} & \textbf{70.32$\pm$4.96} & \textbf{72.06$\pm$5.43} & \textbf{70.32$\pm$4.96} & \textbf{70.38$\pm$4.76} \\
			\hline
	\end{tabular}}
	\caption{Binary classification results using Linear-SVM and MLP classification models. The results are averaged over three random subject splits.}\label{classification_table}
	\vspace{-0.25in}
\end{table*}

\noindent\textbf{Data Preparation:} For both datasets described in Section \ref{section_dataset}, the time-series signals were zero-centered and normalized to have unit standard-deviation. Next, non-overlapping frames of length 250 time-steps were extracted from each time-series signal. Note, the source dataset consists of a single wrist worn accelerometer sensor; whereas the target dataset uses a different accelerometer sensor and consists of six sensors placed at different parts of the body. Thus, the data collection protocol and data distribution is completely different between the two datasets.

\noindent\textbf{Feature Extraction:} Using the pre-trained source autoencoder model we explore two variants to extract latent feature representations for the target dataset. For the first variant we do not constrain the length of latent representations obtained from the encoder block. The length of each latent representations after being vectorized is 48000 (6 sensors $\times$ 250 time-steps $\times$ 32 filters). This is too big to be directly used as input to the MLP classification model. Instead we use Principal Component Analysis (PCA) and bring down the length to a 1600 dimensional feature representation, allowing us to retain 98-99\% of the variance exhibited by the data. The total number of non-overlapping frames in the target dataset was equal to 1786. For this reason we decided to set the number of PCA components to 1600. In the second variant we constrain the size of latent representations by using a global-average-pool layer after the encoder. The length of each feature after being vectorized is 192 (6 sensors $\times$ 32 filters).

We also evaluate the performance of two other baseline methods: (1) A 19-dimensional feature vector consisting of different statistics is calculated over each frame \cite{wang2016statistical}; (2) Original normalized time-series signal. The 19-dimensional feature vector includes \emph{mean}, \emph{variance}, \emph{root-mean-square (RMS)} value of the raw accelerations on each of $X$, $Y$ and $Z$ axes, \emph{pearson correlation coefficients ($\rho$)} between $X$-$Y$, $Y$-$Z$ and $X$-$Z$ time series, \emph{difference between maximum and minimum accelerations} on each axis denoted by $dx, dy, dz$, and $\sqrt{dx^2 + dy^2}$, $\sqrt{dy^2 + dz^2}$, $\sqrt{dx^2 + dz^2}$, $\sqrt{dx^2 + dy^2 + dz^2}$. As for the original time-series signal, vectorizing each frame will result in a 4500 dimensional feature representation (6 sensors $\times$ 250 time-steps $\times$ 3 axis). Here too we use PCA to bring down the feature length to 1,600. For all three feature representations we use the same MLP architecture (described in Section \ref{section_network_architecture}) as our target classification model. 


\noindent\textbf{Evaluation:} 
We randomly select equal number of subjects from each class for the training and test sets. Classifying gait patterns into Parkinson's disease and healthy subjects is a non-trivial problem, especially when working with limited data. Also, gait patterns from the two groups share similar statistical summaries as seen in Table \ref{joint_statistics_info}. Subject-bias was avoided by making sure that data samples from the same subject were not present across the training and test splits. 

The binary classification results averaged over three random subject splits is shown in Table \ref{classification_table}. The table shows the mean$\pm$std values for accuracy, precision, recall and F1-score. In addition to using the MLP model, we also evaluate the above features using a Linear Support Vector Machine (Linear-SVM) classifier. PCA representations of the original time-series signal perform the worst in both classification models. This is followed by the 19-dimensional hand-engineered feature. Both variants of the proposed method do better than the two baseline approaches. The constrained variant of the proposed method has a slightly better average performance than the unconstrained version. However, we also observe a larger standard-deviation. We also notice that the proposed method shows similar classification results on both SVM and MLP classifiers. The choice of source dataset used to train the autoencoder model does affect the proposed method's stability, as seen from the standard deviation values. 

\begin{figure}[t!]
	\centering
	\vspace{0.075in}
	\includegraphics[width=0.925\linewidth]{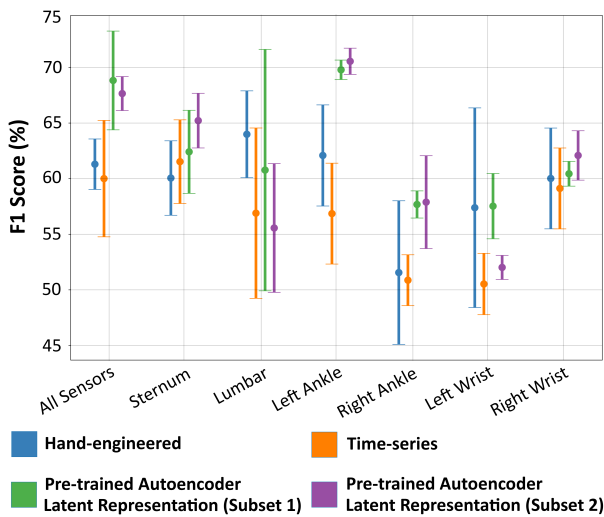}
	\caption{Illustration of the error bar plot of the F1-Score binary classification performance when using different sensors in the target gait dataset. The MLP classification model was used to get these results.}\label{errorbar_plot}
	\vspace{-0.25in}
\end{figure}

The classification results in Table \ref{classification_table} were obtained using all six sensors in the target dataset. Using the MLP classifier, we also examine the influence each feature representation has when using each of the six sensors individually. For this analysis we only consider the unconstrained variant of the proposed method due to its lower standard deviation. Figure \ref{errorbar_plot} displays the error bar information of the F1-Score performance w.r.t. each individual sensor. The \emph{All Sensors} entry in this figure corresponds to MLP classifier's F1-Score entry in Table \ref{classification_table}. Except for \emph{Lumbar} and \emph{Left-Wrist} sensors, the proposed feature representation does comparatively better than the baseline features on all other sensors. The following interesting observations can be made with respect to the \emph{Wrist}, \emph{Ankle} sensors -- (1) the \emph{Left-Ankle} sensor does better than the \emph{Right-Ankle} sensor; (2) the \emph{Right-Wrist} sensor does better than the \emph{Left-Wrist} sensor; (3) the \emph{Left-Ankle} sensor does surprisingly better than the \emph{Right-Wrist} sensor. With regards to the third observation, one would expect to see better results using the \emph{Wrist} sensors since the source dataset consisted of a wrist-worn accelerometer sensor. This could be due to difference in sensor device used and the protocol followed during data collection. 

\vspace{-0.025in}
\section{Conclusion}\label{section_conclusion}
\vspace{-0.025in}

In this paper we explore the use of unsupervised pre-trained autoencoder models to extract feature representations from gait data for Parkinson's disease classification. We trained two different autoencoder models using a larger source dataset comprising of only healthy subjects. We evaluated the impact the choice of source dataset had on the final target (binary) classification task. Our findings indicate that it is indeed possible to adapt models from a very different domain and label-set to another with robust performance.  The source and target datasets used in our experiments came from different data distributions and were collected in different environments. This study opens new possibilities into the use of existing public data-sources of time-series from wearables to learn and adapt features for very specialized low-data use-cases. For instance, in this paper we leveraged data from ADLs, to learn robust features that can be adapted for use in Parkinson's disease assessment, despite both applications having little in common in terms of signal characteristics or class-labels. Possible extensions to this work include: explore the binary-classification ability of pre-trained models under different DBS frequency conditions; use of unsupervised pre-trained models across sensor platforms, like accelerometer to gyroscope and vice versa. 



{\footnotesize
\bibliography{egbib}
\bibliographystyle{ieee}
}
\end{document}